\documentclass[11pt]{article}

\usepackage[utf8]{inputenc}
\usepackage[T1]{fontenc}
\usepackage{lmodern}  
\usepackage{textcomp}  
\usepackage{amsmath,amssymb,amsfonts}
\usepackage{graphicx}
\usepackage[hidelinks]{hyperref}
\usepackage{url}
\usepackage{booktabs}
\usepackage{array}
\usepackage{listings}
\usepackage{xcolor}
\usepackage{algorithm}
\usepackage{algpseudocode}
\usepackage{caption}
\usepackage{subcaption}

\usepackage[margin=1in]{geometry}

\title{MicroSims: A Framework for AI-Generated, Scalable Educational Simulations with Universal Embedding and Adaptive Learning Support}

\author{
    Valerie Lockhart \\
    \texttt{https://www.linkedin.com/in/valockhart/}
    \and
    Dan McCreary \\
    \texttt{https://www.linkedin.com/in/danmccreary/}
    \and
    Troy A. Peterson \\
    \texttt{https://www.linkedin.com/in/troyapeterson/}
}

\date{Version 0.06 \\ \today}

\begin{document}

\maketitle

\begin{abstract}
Educational simulations have long been recognized as powerful tools for enhancing learning outcomes, yet their creation has traditionally required substantial resources and technical expertise. This paper introduces \textit{MicroSims}---a novel framework for creating lightweight, interactive educational simulations that can be rapidly generated using artificial intelligence, universally embedded across digital learning platforms, and easily customized without programming knowledge. MicroSims occupy a unique position at the intersection of three key innovations: (1) standardized design patterns that enable AI-assisted generation, (2) iframe-based architecture that provides universal embedding and sandboxed security, and (3) transparent, modifiable code that supports customization and pedagogical transparency. We present a comprehensive framework encompassing design principles, technical architecture, metadata standards, and development workflows. Drawing on empirical research from physics education studies and meta-analyses across STEM disciplines, we demonstrate that interactive simulations can improve conceptual understanding by up to 30-40\% compared to traditional instruction. MicroSims extend these benefits while addressing persistent barriers of cost, technical complexity, and platform dependence. This work has significant implications for educational equity, and low-cost intelligent interactive textbooks that enabling educators worldwide to create customized, curriculum-aligned simulations on demand. We discuss implementation considerations, present evidence of effectiveness, and outline future directions for AI-powered adaptive learning systems built on the MicroSim foundation.
\end{abstract}

\section{Introduction}

\subsection{The Promise of Interactive Learning}

In Neal Stephenson's visionary novel \textit{The Diamond Age: Or, A Young Lady's Illustrated Primer} \cite{stephenson1995}, he presents a compelling glimpse into the future of education centered around a remarkable interactive book that adapts its narrative and lessons to the specific needs, progress, and circumstances of its reader. While we haven't yet achieved the full sophistication of Stephenson's Primer, today's educational simulations and interactive learning tools are taking significant steps toward this vision of truly adaptive, personalized learning experiences.

Educational research consistently confirms what many intuitively understand: we learn best by doing. Hands-on, experiential learning creates neural pathways that are stronger and more enduring than those formed through passive consumption of information \cite{freeman2014active, prince2004active}. Students retain approximately 75\% of what they learn when they practice by doing, compared to just 5-10\% of what they hear in lectures or read in textbooks. Concepts explored through interactive simulation lead to 30-40\% faster mastery than traditional instructional methods alone \cite{wieman2008phet, rutten2012learning}.

Educational simulations embody this hands-on approach by placing students in interactive environments where they can manipulate variables, observe outcomes, test hypotheses, and develop intuitive understanding through direct experience. This transforms abstract concepts into tangible experiences---making invisible forces visible, compressing time to observe long-term effects, and allowing safe experimentation with potentially dangerous or expensive real-world processes.

\subsection{The Challenge: Barriers to Simulation Adoption}

Despite compelling evidence for their effectiveness \cite{dangelo2014simulations, merchant2014effectiveness}, educational simulations face persistent barriers to widespread adoption:

\begin{itemize}
\item \textbf{Development Costs}: Creating high-quality educational simulations traditionally requires specialized teams of instructional designers, subject matter experts, software developers, and UX designers working through time-intensive development cycles. This makes simulations expensive and limits their availability across the curriculum.

\item \textbf{Technical Complexity}: Traditional simulation platforms often require specific software installations, particular operating systems, or specialized hardware configurations, creating friction for both educators and students.

\item \textbf{Platform Dependence}: Many existing simulations are tightly coupled to specific learning management systems or delivery platforms, limiting their reusability and requiring institutions to adopt particular technological ecosystems.

\item \textbf{Inflexibility}: Once created, traditional educational simulations are often "black boxes"---powerful but difficult to modify. If an educator wants to adjust parameters, add features, or adapt content to match specific curriculum needs, they typically lack the ability to do so.

\item \textbf{Quality Inconsistency}: While some organizations like PhET Interactive Simulations \cite{phet2023} have produced exceptional educational simulations through rigorous research-based development, the broader landscape contains highly variable quality, with many simulations suffering from poor user experience design or pedagogical weaknesses.
\end{itemize}

\subsection{The Opportunity: Generative AI and Standardization}

Recent advances in generative artificial intelligence present unprecedented opportunities to address these barriers. Large language models like GPT-4 and Claude have demonstrated remarkable capabilities in generating functional code from natural language descriptions. However, successfully applying AI to educational simulation development requires more than raw generative capability---it requires carefully designed frameworks, standardized patterns, and best-practice guidelines that AI systems can reliably implement.

\begin{figure}[htbp]
\centering
\includegraphics[width=0.6\textwidth]{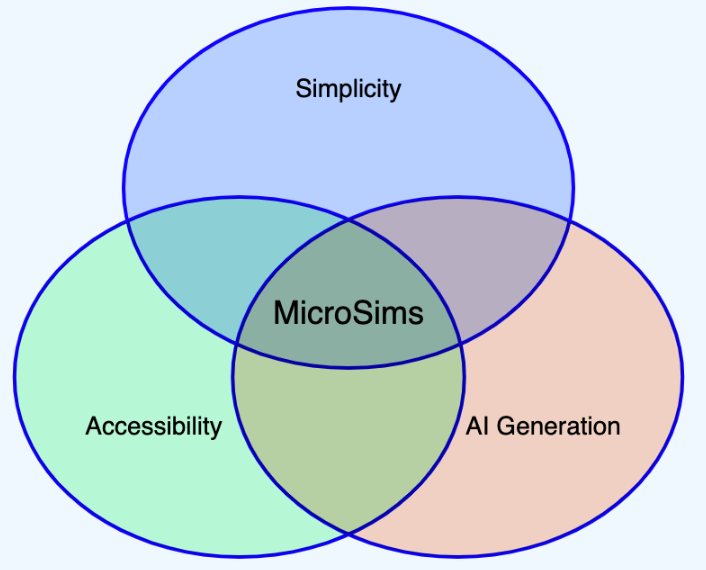}
\caption{MicroSims occupy a unique position at the intersection of three critical characteristics: Simplicity (lightweight, focused scope), Accessibility (browser-based, universal embedding), and AI Generation (standardized patterns, prompt-compatible). This convergence of attributes distinguishes MicroSims from other educational technology approaches and enables their scalable deployment across diverse learning contexts.}
\label{fig:uniqueness}
\end{figure}

This paper introduces \textit{MicroSims}: a comprehensive framework for creating lightweight, interactive educational simulations that occupy a unique position at the intersection of three key characteristics (Figure~\ref{fig:uniqueness}):

\begin{enumerate}
\item \textbf{Simplicity}: Focused simulations with clear parameters, constrained scope, and transparent code that is tractable for AI generation and human modification
\item \textbf{Accessibility}: Universal embedding via iframe architecture, responsive design for multiple devices, and compatibility with any platform supporting basic web standards
\item \textbf{AI Generation}: Standardized design patterns and documented best practices that enable rapid creation and iterative refinement through large language models
\end{enumerate}

A collection of over 100 MicroSim examples demonstrating these principles across diverse subject areas is available at \url{https://dmccreary.github.io/microsims/} \cite{microsims2024}.

\subsection{Research Contributions}

This work makes several distinct contributions to the fields of educational technology and computer-supported collaborative learning:

\begin{enumerate}
\item \textbf{Comprehensive Design Framework}: We present a complete framework for educational simulation design encompassing technical architecture, pedagogical principles, user experience guidelines, and accessibility standards. This framework has been validated through the creation of over 100 MicroSim examples across diverse subject areas.

\item \textbf{AI-Compatible Standardization}: We demonstrate how carefully structured design patterns and coding conventions enable generative AI systems to create pedagogically sound, functionally correct educational simulations from natural language descriptions. This represents a novel approach to leveraging AI for educational content creation.

\item \textbf{Universal Embedding Architecture}: We show how iframe-based distribution combined with width-responsive design enables a single simulation to work seamlessly across learning management systems, interactive textbooks, mobile devices, and other digital learning environments---without requiring platform-specific versions or complex integration protocols.

\item \textbf{Metadata and Discovery System}: We present a metadata framework based on Dublin Core standards specifically tailored for AI-generated educational simulations, supporting discovery, personalization, learning analytics, and integration with adaptive learning systems.

\item \textbf{Development Workflow}: We document a complete workflow for simulation creation, from initial concept through AI-assisted generation, iterative refinement, testing, and deployment, demonstrating how educators without programming expertise can create custom simulations aligned with specific learning objectives.

\item \textbf{Empirical Foundation}: Drawing on extensive research from the PhET project and meta-analyses across STEM education, we provide evidence-based guidelines for simulation effectiveness and discuss how MicroSims extend proven principles while addressing adoption barriers.
\end{enumerate}

\subsection{Paper Organization}

The remainder of this paper is organized as follows: Section \ref{sec:related} positions MicroSims within the broader landscape of educational technology, differentiating them from existing simulation platforms and interactive learning tools. Section \ref{sec:definition} provides a formal definition of MicroSims and their key characteristics. Section \ref{sec:framework} presents the comprehensive design framework encompassing pedagogical principles and technical standards. Section \ref{sec:architecture} details the technical implementation including the p5.js foundation, responsive design patterns, and iframe integration. Section \ref{sec:metadata} describes the metadata schema for discovery and learning analytics. Section \ref{sec:workflow} documents the AI-assisted development workflow. Section \ref{sec:benefits} presents empirical evidence for simulation effectiveness drawing on educational research. Section \ref{sec:discussion} discusses implications, limitations, and broader impact. Section \ref{sec:conclusion} concludes and outlines future research directions.

\section{Related Work}
\label{sec:related}

The educational technology landscape has undergone significant transformation over the past decade, with interactive simulations, virtual laboratories, and adaptive learning platforms becoming increasingly prevalent in formal and informal learning environments. This section positions MicroSims within this broader ecosystem, examining how they differentiate from and complement existing technologies.

\subsection{Traditional Educational Simulations}

Traditional educational simulations, such as PhET Interactive Simulations from the University of Colorado \cite{wieman2008phet, phet2023} or NetLogo models from Northwestern University \cite{wilensky1999netlogo}, represent sophisticated educational tools that have proven effective in science and mathematics education. PhET simulations alone receive over 45 million simulation runs annually, with extensive research demonstrating their effectiveness in improving student understanding of physics concepts \cite{adams2008study, finkelstein2005phet}.

However, these platforms are characterized by several limitations that MicroSims explicitly address:

\textbf{Development Resources}: Traditional simulations typically require significant development resources, specialized programming expertise, and ongoing maintenance to ensure compatibility across evolving web technologies. A single PhET simulation can take 6-12 months and hundreds of person-hours to develop \cite{phet2023}. MicroSims employ standardized architectural patterns that enable automated generation through large language models, reducing development time to minutes or hours.

\textbf{Feature Complexity}: Existing simulation platforms often implement comprehensive feature sets that, while powerful, can overwhelm both educators seeking to integrate specific concepts and students encountering cognitive overload. MicroSims adopt a deliberately constrained approach, focusing on specific learning objectives with minimal extraneous functionality. This constraint-based design philosophy aligns with cognitive load theory principles \cite{sweller1988cognitive}, which suggest that learning is optimized when instructional materials minimize irrelevant cognitive processing.

\textbf{Limited Integration}: Traditional educational simulations frequently operate as standalone applications with limited integration capabilities. MicroSims are architected from the ground up for embedding within larger educational ecosystems, including intelligent textbooks, learning management systems, and adaptive learning platforms.

\subsection{Interactive Textbooks and Digital Learning Materials}

The interactive textbook market has evolved considerably, with platforms such as Pearson MyLab, McGraw-Hill Connect, and Wiley WileyPLUS offering multimedia-enhanced learning experiences. However, these platforms typically employ pre-authored interactive elements that cannot be easily modified or extended by individual educators.

MicroSims fundamentally differ by providing a \textit{generative} approach to interactive content creation, where simulations are produced on-demand to address specific pedagogical requirements. Furthermore, commercial interactive textbook platforms operate under proprietary licensing models that limit institutional flexibility and long-term sustainability. We recommend that MicroSims generate source code licensed under a Creative Commons Sharealike Attribution Non-commercial license that institutions can freely modify, redistribute, and maintain independently.  We request that organizations that reuse MicroSims give the original authors attribution and we suggest that authors retain the copyright to their MicroSims and restrict the ability to resell MicroSims.

\subsection{Learning Management Systems}

Learning Management Systems (LMS) such as Canvas, Blackboard, and Moodle provide comprehensive platforms for course delivery and student management but rely heavily on external content providers for interactive educational materials \cite{lms2023}. MicroSims complement existing LMS infrastructure by providing a standardized method for generating and deploying interactive content directly within these platforms through iframe embedding. The lightweight architecture of MicroSims ensures compatibility across different LMS implementations without requiring platform-specific adaptations.

\subsection{Virtual Laboratory Platforms}

Virtual laboratory platforms, including Labster and Beyond Labz, offer sophisticated simulation environments for science education but typically require subscription-based access and specialized hardware resources. MicroSims provide an alternative approach that prioritizes accessibility and scalability over comprehensive simulation fidelity. While virtual laboratories excel in providing high-fidelity replications of complex scientific processes, MicroSims focus on isolating and illustrating specific conceptual relationships that support understanding of fundamental principles.

\subsection{AI-Generated Educational Content}

Recent work has explored the use of generative AI for creating educational content, including Khan Academy's Khanmigo \cite{khanacademy2023} and various tutoring systems. However, these systems primarily focus on text-based interactions rather than interactive visual simulations. MicroSims represent a novel approach that leverages AI for generating interactive, visual learning experiences rather than conversational tutoring.

\subsection{Gap Analysis}

Despite the proliferation of educational technologies, a significant gap exists for interactive simulations that are simultaneously: (1) simple enough for AI generation, (2) sophisticated enough for meaningful learning, (3) universally embeddable across platforms, (4) easily customizable by educators, and (5) grounded in empirical research on learning effectiveness. MicroSims are specifically designed to fill this gap, providing a framework that addresses these requirements while maintaining pedagogical rigor and technical accessibility.

\section{MicroSims: Definition and Characteristics}
\label{sec:definition}

\subsection{Formal Definition}

We define an \textit{Educational MicroSim} as a lightweight, standalone interactive simulation that executes within standard web browsers and is specifically designed for pedagogical applications. Educational MicroSims are characterized by their focused scope, browser-native implementation, and compatibility with generative AI systems. They occupy a unique position at the intersection of simplicity, accessibility, and AI-generation capability, providing an optimal balance of interactivity and pedagogical effectiveness.

\subsection{Core Characteristics}

\textbf{Focused Scope}: MicroSims deliberately constrain their focus to specific learning objectives rather than attempting comprehensive coverage of broad subject domains.

\textbf{Browser-Native}: All MicroSims run entirely in web browsers using HTML5, CSS, and JavaScript, requiring no installation or specialized software.

\textbf{Generative AI Compatible}: MicroSims follow standardized design patterns that enable large language models to generate, modify, and extend them based on natural language descriptions.

\textbf{Universal Embedding}: Through iframe architecture, MicroSims can be embedded in any digital environment that supports web content.

\textbf{Transparent Implementation}: MicroSim code is intentionally readable and well-documented, enabling educators and students to examine, understand, and modify the underlying logic.

\subsection{Technical Architecture}

MicroSims are implemented as self-contained web applications, typically using JavaScript frameworks such as p5.js, that require no external dependencies or server-side infrastructure. They follow a standardized width-responsive design pattern with distinct regions for visualization (drawing area) and user controls (interaction area). The layout of MicroSims can be expressed in rules files that are used by generative AI systems to ensure consistent implementation patterns across different simulations.

\begin{figure}[htbp]
\centering
\includegraphics[width=0.6\textwidth]{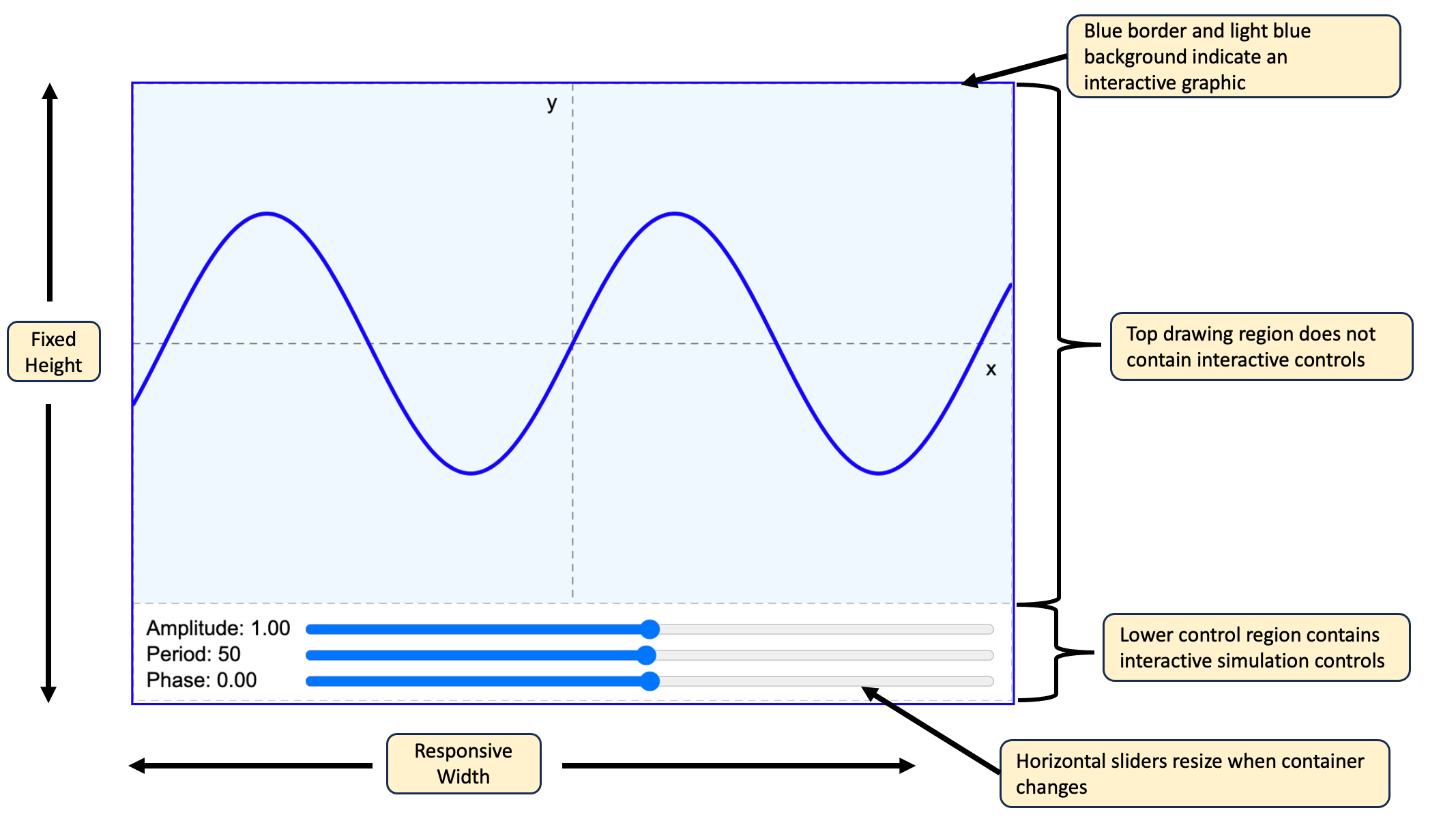}
\caption{MicroSims use a standardized layout pattern with distinct areas for visualization and user controls. This consistent structure facilitates generative AI creation and modification of MicroSims while ensuring usability across diverse educational contexts.  This layout consistency also allows interactivity to be logged in a standardized way for integration with intelligent textbooks and learning analytics systems.}
\label{fig:layout}
\end{figure}

Being browser-based and dependency-free, MicroSims can be easily distributed, embedded in various learning management systems, and accessed across different devices and platforms without installation requirements. The goal is to allow a MicroSim to be placed on any web page using a single HTML \texttt{iframe} element, as illustrated in Figure~\ref{fig:layout}.

\subsection{Educational Purpose and Learning Objectives}

Each MicroSim targets specific learning objectives within a curriculum, enabling students to manipulate parameters and observe resulting changes in real-time. They support experiential learning by allowing learners to explore cause-and-effect relationships through direct interaction with underlying models or algorithms.

The simulations are engineered for modification and extension by non-technical users including educators, students, and content creators. They employ consistent user interface conventions and well-documented code structures to facilitate customization without requiring advanced programming expertise.

MicroSims generate structured event streams capturing user interactions, parameter adjustments, and exploration patterns. These data streams can be analyzed to assess learning progress and provide feedback to adaptive educational systems, including intelligent textbooks that employ reinforcement learning to optimize the learning experience.

\subsection{Scope Boundaries: What MicroSims Are Not}

To clarify the scope and boundaries of Educational MicroSims, it is important to establish what they explicitly are not:

\textbf{Not Simple Animations}: MicroSims are not simple animations of educational concepts. Although generative AI can create beautiful animations, without some student action required for participation, we cannot use feedback and reinforcement learning in intelligent textbooks. Simulations must at a minimum contain interactive controls such as ``Start'' and ``Pause'' buttons. Monitoring user interactions with these controls is critical for the development of intelligent textbooks.

\textbf{Not Technology-Dependent}: MicroSims are not bound to any specific JavaScript library or framework. While our implementation examples utilize p5.js for its pedagogical clarity and ease of use, the MicroSim concept is library-agnostic and can be implemented using vanilla JavaScript, Mermaid.js, ChartJS, Vis-Network.js, Vis-Timeline.js or any other web-based rendering technology that meets the functional requirements.

\textbf{Not Legacy Standards Compliant}: MicroSims do not adhere to traditional e-learning standards such as SCORM (Sharable Content Object Reference Model) or AICC (Aviation Industry Computer-Based Training Committee). These legacy standards impose architectural constraints and complexity that are incompatible with the lightweight, generative nature of MicroSims. However, because MicroSims all have interactive controls, they can be designed to easily work with xAPI standards \cite{xapi2024}, and generative AI can be used to automatically add xAPI calls to the controls area.

\textbf{Not Comprehensive Simulation Environments}: MicroSims are not intended to replace complex, full-featured simulation platforms or virtual laboratories. They are purposefully constrained in scope to address specific, well-defined learning objectives rather than attempting to model entire systems or domains.

\textbf{Not Platform-Specific Applications}: Unlike native mobile applications or desktop software, MicroSims are not tied to specific operating systems or device types. They maintain platform independence through adherence to web standards and responsive design principles.

\textbf{Not Server-Dependent Systems}: MicroSims do not require server-side processing, databases, or cloud infrastructure for their core functionality. While they may optionally integrate with learning analytics platforms, their primary operation remains entirely client-side.

\subsection{Metadata Strategy}

MicroSims leverage established metadata standards where appropriate to enable proper cataloging, discovery, and interoperability within educational repositories and learning management systems. They incorporate Dublin Core metadata elements for resource description, providing standardized fields for title, creator, subject, description, date, type, format, language, and rights information.

This metadata strategy ensures that MicroSims can be systematically organized, searched, and integrated into existing educational technology infrastructures while maintaining their lightweight, generative characteristics. The metadata framework supports both human curation and automated discovery processes, facilitating the scalable deployment of MicroSim collections across diverse educational contexts.

This definition establishes MicroSims as a distinct category of educational technology that bridges the gap between static educational content and complex simulation environments, providing an optimal balance of interactivity, accessibility, and pedagogical effectiveness.

\section{Design Framework}
\label{sec:framework}

The Educational MicroSims Design Framework represents a systematic approach to creating lightweight, interactive educational simulations that prioritize accessibility, responsiveness, and pedagogical effectiveness. This framework establishes standardized design principles, technical architecture patterns, and implementation guidelines that enable both human developers and artificial intelligence systems to create consistent, high-quality educational content.

The design framework emerges from extensive analysis of existing educational simulation platforms and identifies key limitations in current approaches, including complex deployment requirements, inconsistent user interfaces, and limited customization capabilities. By establishing a constraint-based design philosophy that prioritizes simplicity and consistency over feature comprehensiveness, the MicroSims framework enables the creation of educational simulations that are both pedagogically effective and technically sustainable.

\subsection{Design Principles}

\subsubsection{Responsive Architecture}

The foundational principle of the MicroSims design framework is responsive adaptability, specifically engineered for educational contexts where content must be accessible across diverse devices and screen configurations. Unlike traditional responsive web design that adapts both horizontal and vertical dimensions, MicroSims employ a constrained responsive model where simulations maintain fixed heights while dynamically adjusting to container width variations. This approach ensures consistent educational experiences across different devices while maintaining the precise spatial relationships necessary for effective data visualization and interactive elements.

The responsive design implementation utilizes container queries rather than viewport-based media queries, enabling MicroSims to adapt to their embedding context rather than the overall device screen. This architectural decision is particularly critical for educational applications where simulations may be embedded within learning management systems, digital textbooks, or other educational platforms with complex layout structures.

Container width adaptation is implemented through a standardized \texttt{updateCanvasSize()} function that recalculates layout parameters based on the detected container dimensions. This function triggers automatic repositioning of user interface elements, rescaling of text sizes within defined bounds, and adjustment of visualization areas to maintain optimal information density.

\subsubsection{Accessibility by Design}

Accessibility considerations are integrated throughout the MicroSims design framework, ensuring that educational simulations remain usable by learners with diverse abilities and assistive technology requirements. The framework mandates the implementation of semantic HTML structures, appropriate ARIA (Accessible Rich Internet Applications) labeling, and keyboard navigation support for all interactive elements. Color schemes are selected to provide sufficient contrast ratios that meet or exceed WCAG 2.1 AA standards, and information is never conveyed through color alone.

The accessibility framework includes specific provisions for screen reader compatibility, with standardized \texttt{describe()} function implementations that provide comprehensive textual descriptions of simulation content and interactions. These descriptions are automatically generated during the setup phase and dynamically updated as simulation states change, ensuring that users relying on assistive technology receive equivalent information to visual users.

Motor accessibility considerations include minimum touch target sizes of 44 by 44 pixels for mobile interfaces, adequate spacing between interactive elements, and support for alternative input methods. The framework accommodates users who may have difficulty with precise pointer control by implementing forgiving interaction zones and providing alternative interaction methods where appropriate.

In alignment with Universal Design for Learning (UDL), MicroSims provide multiple means of engagement, representation, and expression through multimodal interfaces and adaptable user controls. This framework ensures that diverse learners can access content through their preferred modalities, engage with materials in ways that maintain motivation and interest, and demonstrate understanding through varied interaction patterns. The responsive design and flexible interface elements inherently support UDL principles by enabling personalization and adaptability across different learning contexts and individual needs.

\subsubsection{Standards-Based Development}

The MicroSims framework adheres to web standards and best practices, ensuring long-term compatibility and interoperability across different platforms and technologies. All simulations are implemented using standard HTML5, CSS3, and JavaScript technologies without proprietary extensions or vendor-specific features. This standards-based approach ensures that MicroSims remain functional across different web browsers and can be easily maintained as web technologies evolve.

Dublin Core metadata standards are integrated into the framework to support resource discovery and cataloging within educational repositories. Each MicroSim includes standardized metadata elements describing educational objectives, subject matter, difficulty level, and technical requirements. This metadata is structured using JSON Schema specifications that enable automated validation and processing by educational content management systems.

\subsection{Pedagogical Foundations}

\subsubsection{Learning Objectives Alignment}

The pedagogical foundation of the MicroSims framework is rooted in constructivist learning theory, which emphasizes active knowledge construction through hands-on exploration and experimentation. The framework provides structured approaches for aligning simulation design with specific learning objectives, ensuring that interactive elements directly support intended educational outcomes. This alignment is facilitated through systematic learning objective decomposition, where complex concepts are broken into discrete, explorable components.

The framework incorporates Bloom's Taxonomy as an organizational structure for categorizing learning objectives and selecting appropriate interaction patterns. Lower-level objectives (remembering, understanding) are supported through guided exploration interfaces with clear feedback mechanisms. Higher-level objectives (analyzing, evaluating, creating) are addressed through open-ended parameter spaces that enable hypothesis testing and creative exploration. The progression from structured to open-ended interactions supports scaffolded learning experiences.

Assessment integration is embedded within the pedagogical framework through unobtrusive data collection that captures learning indicators without disrupting the exploration process. Interaction patterns, parameter choices, and exploration sequences provide rich data sources for formative assessment.

\subsubsection{Cognitive Load Management}

The framework explicitly addresses cognitive load theory principles through constrained interface design that minimizes extraneous cognitive processing. Visual design elements follow established principles of multimedia learning, with coordinated presentation of textual and visual information that supports rather than competes for cognitive resources. Color coding, spatial organization, and progressive disclosure techniques are systematically employed to manage information complexity.

Intrinsic cognitive load is managed through careful selection of simulation complexity relative to learner expertise levels. The framework provides guidelines for determining appropriate parameter ranges, interaction granularity, and feedback frequency that match learner capabilities. Extraneous cognitive load is minimized through consistent interface conventions, predictable interaction patterns, and elimination of decorative elements that do not support learning objectives.

Germane cognitive load is optimized through design patterns that encourage schema construction and knowledge transfer. The framework promotes the use of analogies, real-world connections, and cross-simulation consistency that support broader conceptual understanding.

MicroSims intentionally guide learners through a semantic wave---unpacking abstract concepts into tangible, interactive experiences, and then repacking them into generalized understanding. This pedagogical pattern signals that MicroSims don't just entertain---they develop conceptual transfer by moving learners between concrete manipulation and abstract reasoning. The interactive nature of simulations enables this wave-like progression, where students ground abstract principles in observable phenomena before reconstructing more sophisticated conceptual frameworks that transcend specific examples.

\subsubsection{Adaptive Learning Support}

The framework incorporates provisions for adaptive learning experiences that can adjust to individual student needs and preferences. Standardized data collection protocols capture detailed interaction logs that can inform adaptive algorithms about student understanding, engagement levels, and learning preferences. This data enables intelligent tutoring systems to make informed decisions about content sequencing, difficulty adjustment, and intervention timing.

Personalization features include adjustable complexity levels, alternative representation modes, and customizable interface preferences. The framework supports multiple learning modalities through coordinated visual, auditory, and kinesthetic interaction options. The adaptive framework includes provisions for real-time difficulty adjustment based on student performance indicators.

\subsubsection{PRIMM Methodology Integration}

The MicroSims framework brings the principles of constructivist learning theory to life by guiding students through a structured, hands-on progression modeled after the PRIMM methodology---Predict, Run, Investigate, Modify, and Make. In the automatically generated lesson plan for each MicroSim, learners begin by predicting what they think will happen in a dynamic simulation, then run the interactive model to observe real outcomes. They investigate relationships between variables, experiment by modifying parameters using buttons and slider controls, and if they have access to the right tools, they can generate their own versions or extensions of the simulation. This cycle promotes active engagement, conceptual understanding, and creative confidence, helping learners move from passive observation to authentic construction of knowledge through experimentation and iteration. The PRIMM framework aligns naturally with the MicroSims architecture, where transparent code, modifiable parameters, and extensible designs explicitly support this progression from prediction through creation.

\subsection{Implementation Standards and Guidelines}

\subsubsection{Code Architecture and Organization}

The implementation standards define a comprehensive code organization structure that promotes consistency, maintainability, and educational transparency. The standardized architecture separates global variables, setup functions, draw loops, interaction handlers, and utility functions into clearly defined sections with consistent naming conventions. This organization enables educators and students to quickly locate and understand different aspects of simulation functionality.

Variable naming follows educational conventions that prioritize clarity over brevity, with descriptive names that indicate both purpose and units where applicable. The framework mandates comprehensive inline documentation that explains both technical implementation details and pedagogical rationale for design decisions.

Function organization follows a hierarchical structure where high-level educational functions call lower-level technical implementation functions. This structure enables educators to focus on pedagogical customization without requiring deep technical expertise in graphics programming or interaction handling.

\subsubsection{User Interface Design Patterns}

The framework defines comprehensive user interface design patterns that ensure consistent user experiences across different MicroSims while accommodating diverse educational content requirements. Standardized control placement positions all interactive elements within a designated control region below the main visualization area, providing predictable interface layouts that reduce cognitive load associated with navigation and control discovery.

Control element styling follows platform conventions while maintaining educational appropriateness and accessibility compliance. Slider controls use consistent visual styling, labeling patterns, and value display formats. Button interfaces employ standardized sizing, color schemes, and feedback mechanisms.

The framework provides specific guidelines for title positioning, with automatic centering at the top of the canvas area using responsive text sizing algorithms. Labels and instructions follow consistent placement patterns relative to associated controls, with adequate spacing for touch interaction and visual clarity.

\subsubsection{AI Integration and Quality Assurance}

The framework is specifically designed to support automated generation by artificial intelligence systems, particularly large language models capable of code synthesis. The standardized patterns and templates provide clear reference implementations that AI systems can modify and adapt for specific educational requirements. This design consideration enables rapid content creation while maintaining quality and consistency standards.

Comprehensive testing protocols ensure that MicroSims meet educational effectiveness and technical reliability standards across diverse deployment environments. Testing procedures include functionality verification across different browsers and devices, accessibility compliance validation, and educational effectiveness assessment through user studies with target learner populations.

\section{Technical Architecture}
\label{sec:architecture}

The technical architecture of Educational MicroSims represents a carefully balanced approach to creating interactive educational content that maximizes accessibility, maintainability, and generative AI compatibility. The architecture prioritizes web standards, responsive design principles, and educational transparency while minimizing technical complexity and deployment requirements.

The design decisions underlying the MicroSims architecture emerge from extensive analysis of educational technology deployment scenarios, ranging from individual student devices to institutional learning management systems. The architecture addresses the fundamental challenge of creating interactive content that functions reliably across diverse technical environments while remaining simple enough for educators and students to understand, modify, and extend.

\subsection{Technology Stack}

\begin{figure}[htbp]
\centering
\includegraphics[width=0.6\textwidth]{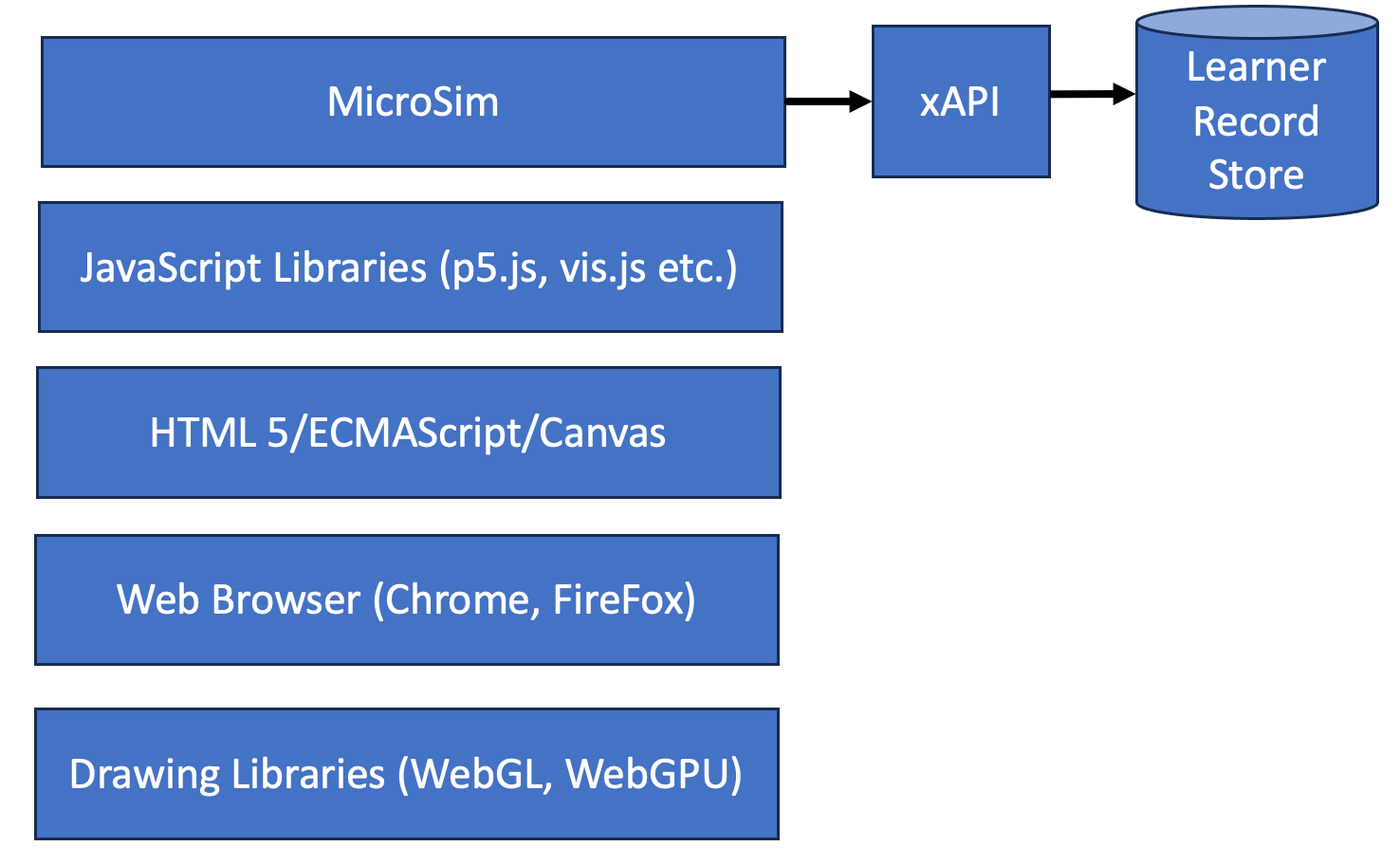}
\caption{MicroSims use a standardized web deployment architecture depicted as a series of layers. The foundation is built on HTML-5, JavaScript standards (ECMAScript), ensuring broad compatibility across browsers and devices. Modern browsers such as Chrome ane FireFox can leverage consistent low-level drawing libraries such as WebGL and WebGPU for fast rendering of complex 3D simulations. The p5.js library provides the core drawing and interaction capabilities, while the MicroSims framework extends p5.js with standardized patterns for responsive design, user interface management, and educational data collection. This layered architecture supports seamless embedding within iframe elements, enabling integration with diverse educational platforms while maintaining security and performance.  A MicroSIm can optionally generate xAPI JSON statements to report user interactions to learning record stores for integration with intelligent textbooks and learning analytics systems.  Analysis of student interaction data can be used to adaptively modify the simulation experience using reinforcement learning techniques.}
\label{fig:architecture}
\end{figure}

\subsubsection{p5.js Foundation}

The technical architecture of the MicroSims framework is built upon the p5.js creative coding library, selected for its educational transparency, extensive documentation, and gentle learning curve. p5.js provides a comprehensive set of drawing and interaction primitives while maintaining code readability that enables educators and students to understand and modify simulation logic. The framework extends p5.js capabilities with standardized patterns for responsive design, user interface management, and educational data collection.

The selection of p5.js as the foundational technology reflects several critical architectural requirements. First, p5.js prioritizes educational accessibility through its simplified syntax and comprehensive documentation ecosystem. Second, the library's focus on immediate visual feedback aligns perfectly with the pedagogical goals of interactive simulations. Third, p5.js maintains broad browser compatibility without requiring complex build processes or development toolchains, enabling direct deployment and modification in educational environments.

The architecture defines a modular structure where core simulation logic is separated from presentation and interaction layers. This separation enables simulations to be easily modified or extended without affecting the underlying educational model. The framework provides standardized templates for common simulation types, incorporating best practices for code organization, naming conventions, and documentation standards.

\subsubsection{MicroSim Rules Files}

A critical component of the MicroSims architecture is the comprehensive rules file system that enables generative AI systems to create consistent, high-quality educational simulations. These rules files codify the design patterns, layout conventions, and implementation standards that ensure uniformity across different AI-generated content while maintaining educational effectiveness.  Modern generative AI systems can utilize these rules files to produce MicroSims that adhere to established pedagogical and technical standards.  Examples of these rules are `skills` files that are mapped to specific content generation tasks.

The rules files define three primary layout types that accommodate different educational simulation requirements. Fixed layouts provide simple positioning for basic demonstrations, responsive width layouts adapt horizontally to container dimensions while maintaining fixed heights, and two-column layouts separate simulation visualization from data analysis components. Each layout type includes specific implementation patterns, variable naming conventions, and responsive behavior specifications.

The standardization extends to user interface design patterns, including consistent control placement, labeling conventions, and interaction feedback mechanisms. All interactive elements follow prescribed positioning relative to the drawing area, with sliders expanding to utilize available width and buttons maintaining consistent spacing and styling. The rules ensure that generated simulations include proper accessibility features, responsive behavior, and educational transparency.

Precise rules files are critical for efficient generation of MicroSims by large language models (LLMs).  By carefully structuring rules files only the relevant rules need to be brought into the context window of the LLM, allowing it to focus on the specific generation task without being overwhelmed by extraneous information.  This targeted approach enhances the quality and consistency of AI-generated simulations while reducing computational overhead.

\subsubsection{HTML5 Canvas and Cross-Platform Compatibility}

The MicroSim framework leverages HTML5 Canvas capabilities to provide rich interactive experiences while maintaining compatibility across diverse devices and platforms. Canvas-based rendering ensures consistent visual output across different browsers and operating systems, while HTML5 audio and video elements support multimedia integration when pedagogically appropriate. The architecture avoids features that require plugin installation or platform-specific implementations, ensuring universal accessibility.

Cross-platform compatibility is validated through systematic testing across major web browsers and mobile platforms. The framework includes performance optimization techniques that ensure smooth operation on lower-powered devices commonly found in educational environments, including older tablets and budget smartphones. Browser compatibility considerations include progressive enhancement strategies where advanced features degrade gracefully on older platforms while maintaining core functionality.

\subsection{Width-Responsive Design Implementation}

Width-responsive design represents the optimal architectural approach for Educational MicroSims, providing essential flexibility without unnecessary complexity. This design philosophy ensures that simulations automatically adjust their horizontal dimensions to match their container while maintaining a fixed height, creating an ideal balance between adaptability and predictability.  An example of a width-responsive rule is to place the title in the top of the canvas and centered at half the width of the canvas.  Using the proper textAlign setting in p5.js makes this easy to implement.

The width-responsive approach addresses critical educational deployment scenarios. In learning management systems, MicroSims fit within content columns of different widths without horizontal scrolling. On mobile devices, students can interact with simulations on phones and tablets with touch controls remaining accessible regardless of screen width. In classroom projection scenarios, teachers benefit from full utilization of projector width with larger controls and text for back-of-room visibility.

The technical implementation involves container detection, where the simulation reads the parent container's width on initialization, dynamic canvas creation with container width and fixed height, and proportional control positioning relative to canvas width. Resize event handling updates the layout when window dimensions change, while content scaling adjusts visual elements proportionally to width changes.

\subsection{Layout Architecture}

\subsubsection{Canvas Regions}

The MicroSims layout architecture follows a standardized structure that divides the canvas into distinct functional regions. This approach ensures visual consistency across different simulations while providing clear separation between interactive content and user controls.

\begin{lstlisting}[caption={Standard MicroSim Canvas Structure}]
// Canvas dimensions
let canvasWidth = 400;              // Initial width (responsive)
let drawHeight = 400;                // Simulation area height
let controlHeight = 50;              // Controls area height
let canvasHeight = drawHeight + controlHeight;
let margin = 25;                     // Margin for visual elements
let sliderLeftMargin = 105;          // Left margin for slider positioning
\end{lstlisting}

The drawing area occupies the upper portion of the canvas with an 'aliceblue' background, providing a visually distinct region for simulation content. The controls area below uses a white background and contains all interactive elements including sliders, buttons, and labels. Both regions are outlined with a 1-pixel silver border to provide clear visual separation.

\subsubsection{Responsive Design Patterns}

The responsive design implementation utilizes container queries rather than viewport-based media queries, enabling MicroSims to adapt to their embedding context rather than the overall device screen. This architectural decision is particularly critical for educational applications where simulations may be embedded within learning management systems, digital textbooks, or other educational platforms with complex layout structures.

Container width adaptation is implemented through a standardized \texttt{updateCanvasSize()} function that recalculates layout parameters based on the detected container dimensions. This function triggers automatic repositioning of user interface elements, rescaling of text sizes within defined bounds, and adjustment of visualization areas to maintain optimal information density. The responsive system includes provisions for progressive disclosure, where complex simulations may hide or simplify certain interface elements on narrow displays.

\subsection{iframe Integration}

\subsubsection{Embedding Protocol}

A critical architectural requirement is seamless integration within iframe elements, enabling MicroSims to be embedded as single-line HTML elements within diverse educational platforms. The framework addresses common iframe integration challenges, including cross-origin communication, responsive sizing, and event handling. MicroSims are designed to function completely within iframe boundaries without requiring parent page modification or cross-frame scripting.

The iframe integration model supports both static and dynamic embedding scenarios. Static embedding involves simple HTML iframe tags with specified dimensions, suitable for content management systems and learning management platforms. Dynamic embedding utilizes JavaScript-based integration that can automatically adjust iframe dimensions based on content requirements and respond to container size changes.

The standardized embedding approach ensures that educational platforms can integrate MicroSims using minimal HTML code while maintaining full functionality. The responsive design automatically adapts to iframe dimensions, ensuring that simulations remain usable regardless of the embedding context or container constraints.

\subsubsection{Security Considerations}

Security considerations for iframe deployment include content security policy compliance and prevention of clickjacking vulnerabilities. The framework implements appropriate sandbox attributes and communication protocols that enable safe embedding while maintaining necessary functionality. Cross-origin resource sharing (CORS) policies are configured to support integration across different domain environments commonly found in educational technology ecosystems.

The architecture prioritizes client-side execution to minimize security risks associated with server-side processing or external data dependencies. All simulation logic executes within the browser environment, reducing potential attack vectors while ensuring that educational content remains accessible even in restrictive network environments.

Data privacy considerations are embedded throughout the architecture, with optional learning analytics integration designed to comply with educational privacy requirements. The framework provides clear separation between core simulation functionality and data collection capabilities, enabling institutions to deploy MicroSims in compliance with their specific privacy policies and regulatory requirements.

\subsubsection{Extended JavaScript Ecosystem}

While p5.js provides the foundational capabilities for most educational simulations, the MicroSims architecture accommodates integration with specialized JavaScript libraries for specific educational domains. For complex network visualizations, the framework supports integration with vis.js and vis-network.js libraries, enabling the creation of interactive graph-based educational content such as concept maps, social network analysis, and algorithm visualization.

Timeline-based educational content leverages vis-timeline.js for creating interactive historical timelines, project scheduling demonstrations, and temporal data analysis simulations. These integrations maintain the core MicroSims design principles while extending capabilities for specific educational requirements that exceed p5.js native functionality.

The extended ecosystem integration follows careful dependency management principles to preserve the lightweight characteristics essential for educational deployment. Each additional library is evaluated for educational necessity, performance impact, and compatibility with the core responsive design requirements.

\section{Metadata and Discovery Framework}
\label{sec:metadata}

As educational technology continues to evolve toward AI-generated, personalized learning experiences, the challenge of organizing and discovering appropriate digital learning resources has become increasingly complex. Educational MicroSims represent a promising approach to personalized, engaging instruction, but as collections of these resources grow into the tens of thousands, educators and institutions need robust systems for cataloging, searching, and integrating these materials into their curricula.

This section presents a comprehensive metadata framework specifically designed for Educational MicroSims that addresses organizational challenges while supporting the pedagogical needs of diverse educational contexts. The framework combines established cataloging standards with education-specific metadata and detailed technical specifications to enable sophisticated search, recommendation, and integration capabilities.

\subsection{Search and Reuse: The Foundation of Educational Resource Discovery}

The fundamental principle underlying effective educational resource management is simple yet critical: educators cannot reuse what they cannot find. Traditional educational resource repositories often suffer from inconsistent cataloging practices, making it difficult for educators to locate materials that match their specific needs. A mathematics teacher seeking a simulation for teaching quadratic functions might struggle to locate appropriate resources among thousands of available options, particularly when materials are tagged inconsistently or lack detailed descriptions of their educational purpose, technical requirements, or pedagogical applications.

The proliferation of AI-generated educational content exacerbates this challenge. While artificial intelligence can rapidly create customized learning materials, these resources require systematic organization to be truly useful at scale. Without standardized metadata, even the most sophisticated educational simulation becomes effectively invisible to educators who could benefit from its use. The power of generative AI in creating educational content can only be fully realized when paired with comprehensive metadata frameworks that enable discovery and integration.

Faceted search capabilities represent a particularly powerful approach to educational resource discovery. Rather than relying on simple keyword matching, faceted search enables educators to filter resources across multiple dimensions simultaneously. An educator can specify grade level (9-12), subject area (chemistry), topic (molecular bonding), desired cognitive level (apply or analyze), and technical requirements (tablet compatibility) to receive a precisely curated list of relevant simulations. This approach transforms resource discovery from a time-consuming challenge into an efficient, targeted activity.

\subsection{Metadata Requirements}

The MicroSims metadata framework employs a layered approach that builds upon established standards while incorporating domain-specific requirements for educational technology. This structured approach ensures compatibility with existing systems while providing the detailed specifications necessary for educational applications.

\subsubsection{Dublin Core Foundation}

At its foundation, the metadata schema incorporates Dublin Core metadata standards---internationally recognized elements for describing digital resources. This ensures compatibility with existing educational repositories and library systems while providing essential information for resource management and discovery. The Dublin Core elements integrated into the MicroSims framework include:

\textbf{Core Elements}: Title, creator, subject, description, publisher, date, type, format, identifier, language, relation, coverage, and rights provide the fundamental cataloging information required for institutional repositories and library systems.

\textbf{Educational Adaptation}: While maintaining Dublin Core compliance, the schema extends these elements to address educational-specific requirements. The \textit{subject} field employs controlled vocabularies that prevent confusion caused by synonym variations, ensuring that resources tagged as "mathematics" and "math" appear together in search results. The \textit{type} field uses educational resource type specifications that distinguish between simulations, demonstrations, and interactive assessments.

\textbf{Rights and Licensing}: The Dublin Core rights element is enhanced to support Creative Commons licensing and educational use restrictions, enabling institutions to filter resources based on permissible usage scenarios and compliance requirements.

\subsubsection{Educational Extensions}

The educational metadata component extends beyond basic cataloging to capture pedagogically relevant information essential for curriculum integration and instructional design. These extensions address the specific needs of educators selecting resources for particular learning objectives and contexts.

\textbf{Grade Level Specifications}: Standardized grade level categories from kindergarten through graduate study enable precise targeting of age-appropriate materials. The framework accommodates both traditional grade levels (K-12) and post-secondary categories (Undergraduate, Graduate, Adult Education) to support diverse educational contexts.

\textbf{Learning Objectives and Taxonomy}: Perhaps most significantly, the schema incorporates Bloom's Taxonomy classifications, allowing educators to search for resources that target specific cognitive skill levels. This enables teachers to locate simulations that support factual recall (Remember level), conceptual understanding (Understand level), or creative application (Create level) depending on their instructional goals.

\textbf{Curriculum Standards Alignment}: The framework supports alignment with multiple curriculum standards frameworks, including Common Core State Standards (CCSS), Next Generation Science Standards (NGSS), and International Society for Technology in Education (ISTE) standards. This alignment enables curriculum coordinators to ensure that selected resources support mandated learning standards.

\textbf{Cognitive Load Assessment}: The schema incorporates cognitive load theory principles through structured assessment of intrinsic, extraneous, and germane cognitive load. This information helps educators select appropriate resources based on student cognitive capacity and instructional design principles.

\subsubsection{Technical Specifications}

Technical metadata addresses the implementation requirements essential for successful deployment in diverse technological environments. These specifications enable technical staff to quickly assess integration requirements while supporting educational decision-making about device compatibility and accessibility features.

\textbf{Platform Compatibility}: Canvas dimensions, framework dependencies, browser compatibility, and device requirements provide comprehensive technical specifications. Performance metrics including target frame rates, memory usage, and computational complexity enable informed deployment decisions for institutions with varied technological infrastructure.

\textbf{Accessibility Compliance}: Accessibility features are explicitly documented, supporting inclusive design principles and compliance with educational accessibility standards such as Section 508 and Web Content Accessibility Guidelines (WCAG). This documentation includes screen reader compatibility, keyboard navigation support, and alternative input methods.

\textbf{Responsive Design Documentation}: The framework documents responsive behavior patterns, including breakpoint specifications and adaptive interface elements. This information supports deployment across diverse device ecosystems commonly found in educational environments.

\subsubsection{Simulation Model Documentation}

A unique aspect of the metadata framework involves comprehensive documentation of the underlying simulation models. Unlike black-box educational software, MicroSims benefit from transparent documentation of their mathematical equations, algorithms, assumptions, and limitations. This transparency serves multiple educational purposes and enables informed pedagogical application.

\textbf{Mathematical Foundations}: The schema documents the mathematical equations, algorithms, and computational methods underlying each simulation. This information enables educators to understand exactly what concepts the simulation demonstrates and helps students appreciate the relationship between mathematical models and real-world phenomena.

\textbf{Model Limitations and Assumptions}: Explicit documentation of simplifying assumptions and model limitations provides educational context for simulation results. For example, a physics simulation modeling projectile motion documents not only the kinematic equations used but also simplifying assumptions such as the absence of air resistance.

\textbf{Variable Specifications}: Comprehensive documentation of input variables, parameters, and output measures supports both educational implementation and learning analytics integration. This documentation includes variable types (input, output, intermediate, constant), data types, units of measurement, and acceptable value ranges.

\subsection{Discovery and Cataloging}

When applied consistently across thousands of MicroSims, this metadata framework enables sophisticated search and recommendation capabilities that transform how educators discover and integrate educational resources. The structured approach to resource description supports multiple discovery modalities while enabling automated personalization and curriculum integration.

\textbf{Faceted Search Implementation}: The comprehensive metadata structure enables sophisticated faceted search interfaces where educators can filter resources across multiple dimensions simultaneously. Search interfaces can provide filtering options for grade level, subject area, learning objectives, technical requirements, accessibility features, and pedagogical approaches. This multi-dimensional filtering approach significantly reduces the time required to locate appropriate resources while ensuring pedagogical alignment.

\textbf{Automated Recommendation Systems}: The standardized metadata enables intelligent recommendation systems that can suggest resources based on curricular context, student performance data, and educational objectives. By documenting how learning objectives, difficulty levels, and prerequisite knowledge are structured, the framework enables adaptive learning systems to make evidence-based recommendations about which simulations will best support individual student needs.

\textbf{Curriculum Integration}: The framework supports automated curriculum mapping where educational resources can be systematically aligned with instructional sequences and learning progressions. Standards alignment metadata enables curriculum coordinators to ensure comprehensive coverage of mandated learning objectives while identifying gaps in available resources.

\textbf{Quality Assurance and Curation}: Structured metadata enables automated quality assessment based on completeness of documentation, alignment with educational standards, and technical compliance requirements. This systematic approach to quality assurance supports scalable curation processes essential for large-scale educational resource collections.

\subsection{Learning Analytics Integration}

The metadata framework incorporates comprehensive specifications for learning analytics data collection and analysis, enabling the development of sophisticated educational measurement and adaptive learning capabilities. This integration addresses both the technical requirements for data collection and the educational considerations necessary for meaningful learning assessment.

\textbf{Event Tracking Specifications}: The schema documents standardized event types and data collection protocols that enable consistent learning analytics across different MicroSims. Events are categorized by type (interaction, navigation, learning, performance, engagement, error) and importance level, enabling prioritized data collection that balances analytical value with privacy considerations.

\textbf{Learning Indicators Documentation}: The framework specifies behavioral indicators that provide evidence of learning progress, including both direct measures (correct responses, task completion) and indirect measures (engagement patterns, exploration behaviors). This documentation enables learning analytics systems to make informed inferences about student understanding and skill development.

\textbf{Privacy and Compliance Framework}: Educational data privacy considerations are explicitly addressed through comprehensive documentation of data collection practices, retention policies, and compliance requirements. The framework supports adherence to educational privacy standards including FERPA, GDPR, and institutional data governance policies.

\textbf{Adaptive Learning Support}: The metadata documents which elements of each simulation can be adapted based on student performance and how adaptation algorithms should interpret learning analytics data. This specification enables intelligent tutoring systems to provide personalized learning experiences that adjust difficulty, provide additional scaffolding, or recommend alternative resources based on individual student needs.

\subsection{JSON Schema Structure and Implementation}

The technical implementation of the metadata framework employs JSON Schema specifications that enable both human curation and automated generation of resource descriptions. This structured approach provides the precision necessary for machine processing while maintaining accessibility for human editors and educational practitioners.

\textbf{Hierarchical Organization}: The schema employs a hierarchical structure that groups related metadata elements while maintaining clear separation between different types of information. Core sections include Dublin Core elements, search and discovery metadata, educational specifications, technical requirements, user interface documentation, simulation models, and analytics specifications.

\textbf{Validation and Consistency}: JSON Schema validation rules ensure metadata consistency and completeness across large collections of educational resources. Required fields, enumerated values, and format constraints prevent common metadata quality issues while enabling automated validation workflows that scale to thousands of resources.

\textbf{Extensibility and Versioning}: The schema structure accommodates future extensions while maintaining backward compatibility with existing metadata. Versioning protocols ensure that metadata evolution can occur without disrupting existing educational technology systems or invalidating previously created resource descriptions.

\textbf{AI-Generated Metadata}: The structured schema design specifically supports automated metadata generation by large language models and other AI systems. Clear field definitions, enumerated values, and validation constraints enable AI systems to generate comprehensive, accurate metadata that meets educational and technical requirements without extensive human review.

\textbf{Implementation Example}: The bouncing ball physics simulation demonstrates the framework's comprehensive approach to metadata documentation. The educational metadata specifies grade levels (6-10), subject areas (Physics, Mathematics, Science), and Bloom's taxonomy levels (Understand, Apply, Analyze). Technical specifications document p5.js framework requirements, responsive canvas dimensions (400X430 pixels), and browser compatibility. User interface documentation details the speed control slider with range 0-20, default value 3, and responsive sizing behavior. The simulation model section documents the kinematic equations, collision detection algorithms, and model assumptions including perfect elastic collisions and absence of friction. This comprehensive documentation enables educators to quickly assess the simulation's appropriateness for their instructional context while providing technical staff with implementation requirements and learning analytics systems with standardized data collection specifications.

The metadata framework represents a foundational component for scalable educational technology ecosystems. By providing comprehensive, structured descriptions of educational resources, the framework enables the sophisticated discovery, integration, and analytical capabilities necessary to realize the full potential of digital learning environments. As educational institutions continue to invest in interactive learning resources, systematic metadata frameworks ensure these investments achieve maximum pedagogical impact through enhanced discoverability, appropriate application, and seamless integration into existing curricula.

\section{Development Workflow}
\label{sec:workflow}

The development of Educational MicroSims follows a systematic workflow that leverages existing resources, generative AI, prompt engineering, pre-defined rules such as skills and structured metadata to allow non-programmers to create personalized learning experiences. This workflow transforms educational needs into interactive simulations through a sophisticated pipeline that maintains pedagogical rigor while enabling rapid development and deployment at scale.

The workflow addresses the scalability challenges in educational technology by providing a systematic approach that begins with repository discovery and continues through AI-assisted generation, iterative refinement, comprehensive testing, metadata creation, and intelligent system integration. This process enables educators to create highly targeted simulations that adapt to individual student needs while maintaining technical quality and educational effectiveness.

\subsection{AI-Assisted Generation}

The AI-assisted generation process represents the core innovation of the MicroSims development workflow, enabling educators without programming expertise to create sophisticated interactive simulations through natural language specifications and template-based development patterns.

\subsubsection{Natural Language Specification}

The development process begins with educators providing natural language descriptions of their educational requirements, which are then systematically converted into technical specifications through structured prompting protocols that include references to rules files and skills. This approach enables domain experts to focus on pedagogical considerations while AI systems handle the technical implementation details.

\textbf{Educational Requirements Specification}: Educators provide structured descriptions that include subject area, grade level, learning objectives, duration requirements, and Bloom's taxonomy levels. For example, a specification might request "Use the MicroSim skill to create a new MicroSim for teaching quadratic functions to high school algebra students (grades 9-10) that enables students to explore the relationships between coefficients and parabola characteristics through interactive parameter manipulation."

\textbf{Technical Requirements Translation}: The natural language specifications are systematically translated into technical requirements including framework selection (p5.js), responsive design parameters, control interface specifications, and accessibility compliance requirements. This translation process ensures that educational intentions are preserved while meeting technical implementation standards.

\textbf{Pedagogical Pattern Recognition}: AI systems analyze the educational requirements to identify appropriate pedagogical patterns from the established MicroSim design vocabulary. These patterns include exploration-based interfaces, parameter manipulation controls, real-time feedback mechanisms, and assessment integration points that align with specified learning objectives.

\subsubsection{Pattern-Based Code Generation}

The code generation process employs established design patterns and template structures that ensure consistency, accessibility, and educational effectiveness across different MicroSims. This pattern-based approach enables reliable generation while maintaining the flexibility necessary for diverse educational applications.

\textbf{Template Selection and Adaptation}: MicroSim rules can suggest an appropriate templates based on educational requirements and technical specifications. Templates provide proven interaction patterns, responsive design frameworks, effective control placement, and accessibility considerations that serve as foundations for new simulations.

\textbf{Standardized Architecture Implementation}: Generated code follows the established MicroSim architecture with separated drawing and control regions, standardized variable naming conventions, comprehensive documentation, and consistent user interface patterns. This architectural consistency enables predictable behavior and simplified maintenance across large collections of simulations.

\textbf{Educational Model Integration}: The generation process incorporates the mathematical models, algorithms, and computational methods specified in the educational requirements. Model documentation includes equations, assumptions, limitations, and variable specifications that support both educational transparency and learning analytics integration.

\subsubsection{Iterative Refinement}

The development workflow includes systematic iterative refinement cycles that enable continuous improvement through preview testing, feedback collection, and targeted modifications. This iterative approach ensures that generated simulations meet both educational objectives and usability requirements.

In our experience, generative AI programs struggle with precise placement of graphics and user interface controls. Therefore, iterative refinement is essential to achieve the desired quality and functionality.  This placement can be made more precise through careful design of placement rules that also integrate width-responsive design principles.

\textbf{Preview and Testing Integration}: Modern AI-assisted development environments provide real-time preview capabilities that allow educators to test functionality immediately upon generation. This rapid feedback cycle enables quick identification of areas requiring modification or enhancement without extensive technical review processes.

It is our experience that many simple MicroSims can be created in a single iteration of robust rules files are used.  We have seen many examples where even MicroSims of up to 500 lines can be done is a single pass will little to now modification required.  As more controls are added and more complex graphics are used, multiple iterations may be required.

\textbf{Refinement Prompt Engineering}: Educators can request specific modifications through structured refinement prompts that specify control additions, removals, or modifications while maintaining the underlying architectural patterns. These prompts enable precise adjustments without requiring comprehensive regeneration of the entire simulation.

\subsection{Repository Discovery and Template Selection}

The development workflow begins with systematic discovery of existing educational simulation repositories and identification of appropriate templates that align with educational objectives. This discovery process leverages faceted search capabilities to efficiently locate relevant resources and development starting points.

\textbf{Repository Exploration}: Educators begin by exploring established repositories such as PhET Interactive Simulations from the University of Colorado Boulder, GitHub repositories containing AI-generated microsimulations, and commercial platforms like Gizmos by ExploreLearning. These repositories serve as starting points for discovering existing resources and identifying templates for creating new MicroSims.

\textbf{Faceted Search Implementation}: Modern educational repositories implement sophisticated search capabilities that enable multi-dimensional filtering across subject areas (Mathematics, Science, Computer Science, Engineering), grade levels (Elementary, Middle School, High School, Undergraduate), learning objectives (Bloom's taxonomy levels), and technical requirements (duration, device compatibility, accessibility features).

\textbf{Template Selection Process}: Once educators identify similar MicroSims that align with their needs, these resources serve as templates for generating customized versions. The template selection process involves identifying core functionality, analyzing technical components, and extracting educational elements that can be adapted for new learning contexts.

\subsection{Customization and Modification}

The customization phase enables educators to adapt existing templates and generated simulations to meet specific pedagogical requirements while maintaining technical quality and educational effectiveness.

\subsubsection{Educator-Driven Adaptation}

Educator-driven adaptation processes enable domain experts to modify simulations without requiring extensive programming knowledge while ensuring that educational objectives remain central to the development process.

\textbf{Structured Customization Prompts}: Educators can request specific modifications through comprehensive prompts that specify educational requirements, technical modifications, and pedagogical considerations. These prompts include subject area specifications, learning objective alignments, control interface requirements, visual design preferences, and assessment integration needs.

\textbf{Parameter Modification}: The workflow supports systematic modification of simulation parameters including variable ranges, default values, step increments, and units of measurement. These modifications enable fine-tuning of educational experiences to match specific curriculum requirements and student ability levels.

\textbf{Interface Customization}: Educators can request additions, removals, or modifications to user interface elements including sliders, buttons, checkboxes, dropdown menus, and display components. The customization process maintains responsive design principles and accessibility compliance while accommodating diverse pedagogical approaches.

\subsubsection{Student Exploration and Extension}

The framework supports student-driven exploration and extension activities that enable learners to modify and extend simulations as part of their educational experience.

\textbf{Guided Modification Activities}: Students can participate in structured activities where they modify simulation parameters, add new features, or extend existing functionality under educator guidance. These activities provide authentic programming experiences while reinforcing subject matter learning.

\textbf{Open-Ended Exploration}: Advanced students can engage in open-ended exploration where they identify limitations in existing simulations and propose enhancements that address real-world applications or extend the mathematical models to more complex scenarios.

\subsection{Quality Assurance and Validation}

The workflow incorporates comprehensive quality assurance procedures that ensure generated simulations meet educational effectiveness, technical reliability, and accessibility standards across diverse deployment environments.

\subsubsection{Functional Testing}

Functional testing procedures verify that generated simulations operate correctly across different browsers, devices, and usage scenarios while maintaining performance standards appropriate for educational environments.

\textbf{Cross-Platform Compatibility}: Testing protocols verify functionality across major web browsers (Chrome, Firefox, Safari, Edge) and mobile platforms (iOS Safari, Android Chrome). Performance optimization ensures smooth operation on lower-powered devices commonly found in educational environments, including older tablets and budget smartphones.

\textbf{Responsive Design Validation}: Testing procedures verify that responsive design implementations adapt correctly to different screen sizes and container dimensions. This validation ensures that simulations remain usable when embedded in learning management systems, digital textbooks, or other educational platforms with varying layout constraints.

\textbf{Integration Testing}: Comprehensive testing verifies iframe integration capabilities, learning management system compatibility, and educational analytics data collection functionality. This testing ensures reliable deployment across diverse educational technology ecosystems.

\subsubsection{Pedagogical Review}

Pedagogical review processes ensure that generated simulations effectively support intended learning objectives while maintaining educational best practices and theoretical foundations.

\textbf{Learning Objective Alignment}: Review procedures verify that simulation features, interaction patterns, and assessment opportunities directly support specified learning objectives. This alignment assessment ensures that technical capabilities serve educational purposes rather than existing as standalone features.

\textbf{Cognitive Load Assessment}: Reviews evaluate the cognitive load imposed by simulation interfaces and interaction requirements, ensuring that extraneous cognitive load is minimized while germane cognitive load supporting learning is optimized. This assessment draws on established cognitive load theory principles and educational psychology research.

\textbf{Educational Effectiveness Validation}: Systematic evaluation with representative student populations verifies that simulations achieve intended learning outcomes and that interaction patterns support rather than hinder educational goals. This validation provides evidence-based assessment of pedagogical effectiveness.

\subsubsection{Accessibility Validation}

Accessibility validation ensures that generated simulations comply with educational accessibility standards and provide inclusive learning experiences for students with diverse abilities and assistive technology requirements.

\textbf{Technical Compliance Testing}: Automated and manual testing procedures verify compliance with Web Content Accessibility Guidelines (WCAG 2.1 AA standards), Section 508 requirements, and educational accessibility best practices. This testing includes screen reader compatibility, keyboard navigation functionality, and color contrast verification.

\textbf{Assistive Technology Testing}: Testing procedures verify compatibility with common assistive technologies including screen readers, alternative input devices, and mobility assistance tools. This testing ensures that students with disabilities can access and interact with simulations effectively.

\textbf{Universal Design Validation}: Reviews assess how well simulations implement universal design principles that benefit all learners, including clear visual hierarchies, consistent interaction patterns, and multiple representation modalities that support diverse learning preferences and abilities.

\subsection{Metadata Generation and Documentation}

Once simulations are finalized and validated, the workflow includes systematic generation of comprehensive metadata that enables discovery, cataloging, and integration with educational technology systems. This metadata follows established schema specifications and supports sophisticated search and recommendation capabilities.

\textbf{Automated Metadata Creation}: The workflow leverages AI systems to generate comprehensive JSON metadata files that follow the Educational MicroSim Metadata Schema. This automated process ensures consistency and completeness while reducing the administrative burden on educators. Generated metadata includes Dublin Core elements, educational specifications, technical requirements, user interface documentation, simulation models, and analytics specifications.

\textbf{Educational Context Documentation}: The metadata generation process captures detailed educational context including learning objectives, curriculum standards alignment, prerequisite knowledge requirements, assessment opportunities, and instructional strategy recommendations. This documentation enables sophisticated search and filtering capabilities that help educators locate resources that precisely match their pedagogical needs.

\textbf{Technical Specification Recording}: Comprehensive technical metadata documents framework dependencies, performance characteristics, device compatibility requirements, and accessibility features. This technical documentation enables system administrators and educational technologists to make informed decisions about deployment and integration requirements.

\textbf{Usage Pattern Documentation}: The metadata includes recommendations for pedagogical implementation, suggested classroom activities, assessment questions aligned with different Bloom's taxonomy levels, and extension activities. This usage documentation helps educators understand how to effectively integrate simulations into their instructional practice.

\subsection{Deployment and Integration}

The final phase of the development workflow involves systematic deployment procedures and integration with intelligent learning systems that enable personalized educational experiences and continuous improvement through learning analytics.

\textbf{Learning Management System Integration}: The deployment process includes standardized procedures for integrating MicroSims with common learning management systems through iframe embedding, single sign-on authentication, and grade passback functionality. This integration enables seamless incorporation of simulations into existing course structures and assessment workflows.

\textbf{Intelligent Textbook Integration}: Simulations are integrated into adaptive learning sequences within intelligent textbook systems that provide personalized learning paths, prerequisite assessment, adaptive difficulty adjustment, and remediation triggering based on individual student performance data. This integration transforms isolated simulations into components of coherent, personalized learning experiences.

\textbf{Learning Analytics Implementation}: The deployment process includes implementation of comprehensive learning analytics data collection following xAPI specifications. These analytics capture detailed interaction events, performance measures, engagement indicators, and learning progress markers that enable continuous system improvement and personalized recommendation generation.

\textbf{Recommendation Engine Integration}: Deployed simulations become part of recommendation systems that use collected learning analytics data to make evidence-based suggestions about resource selection, optimal timing for introduction, and personalized parameter settings. These recommendation capabilities enable increasingly sophisticated educational personalization as the system accumulates data from student interactions.

The development workflow represents a comprehensive pipeline that transforms educational needs into personalized, interactive learning experiences through systematic application of AI-assisted generation, iterative refinement, comprehensive validation, structured metadata creation, and intelligent system integration. This workflow addresses the scalability challenges in educational technology while maintaining pedagogical rigor and technical quality, creating a foundation for educational technology ecosystems that become more effective over time through accumulated learning analytics and continuous improvement processes.

\section{Expected Benefits, Limitations and Examples of Intelligent Textbook Integration}
\label{sec:benefits}
\label{sec:limitations}

\subsection{Introduction: Research Foundation for Interactive Simulations}

Educational research consistently demonstrates the effectiveness of interactive simulations for enhancing learning outcomes across STEM disciplines. Meta-analyses and systematic reviews provide compelling evidence that well-designed simulations significantly improve student engagement, conceptual understanding, and long-term knowledge retention \cite{wieman2008phet, rutten2012learning, dangelo2014simulations}. This substantial body of research, accumulated over decades of classroom implementation and controlled studies, establishes interactive simulations as transformative pedagogical tools that address persistent challenges in STEM education.

Studies consistently report that students using interactive simulations exhibit higher interest and participation in STEM lessons, with measurable improvements in both immediate learning outcomes and sustained understanding. Meta-analyses across STEM disciplines demonstrate that students using interactive simulations consistently outperform peers receiving traditional instruction alone:

\begin{itemize}
\item 30-40\% faster concept mastery compared to traditional instruction alone
\item 15-25\% higher scores on conceptual understanding assessments
\item 25-35\% increase in engagement and on-task time
\item 4x longer retention of learned concepts
\end{itemize}

\begin{figure}[htbp]
\centering
\includegraphics[width=0.95\textwidth]{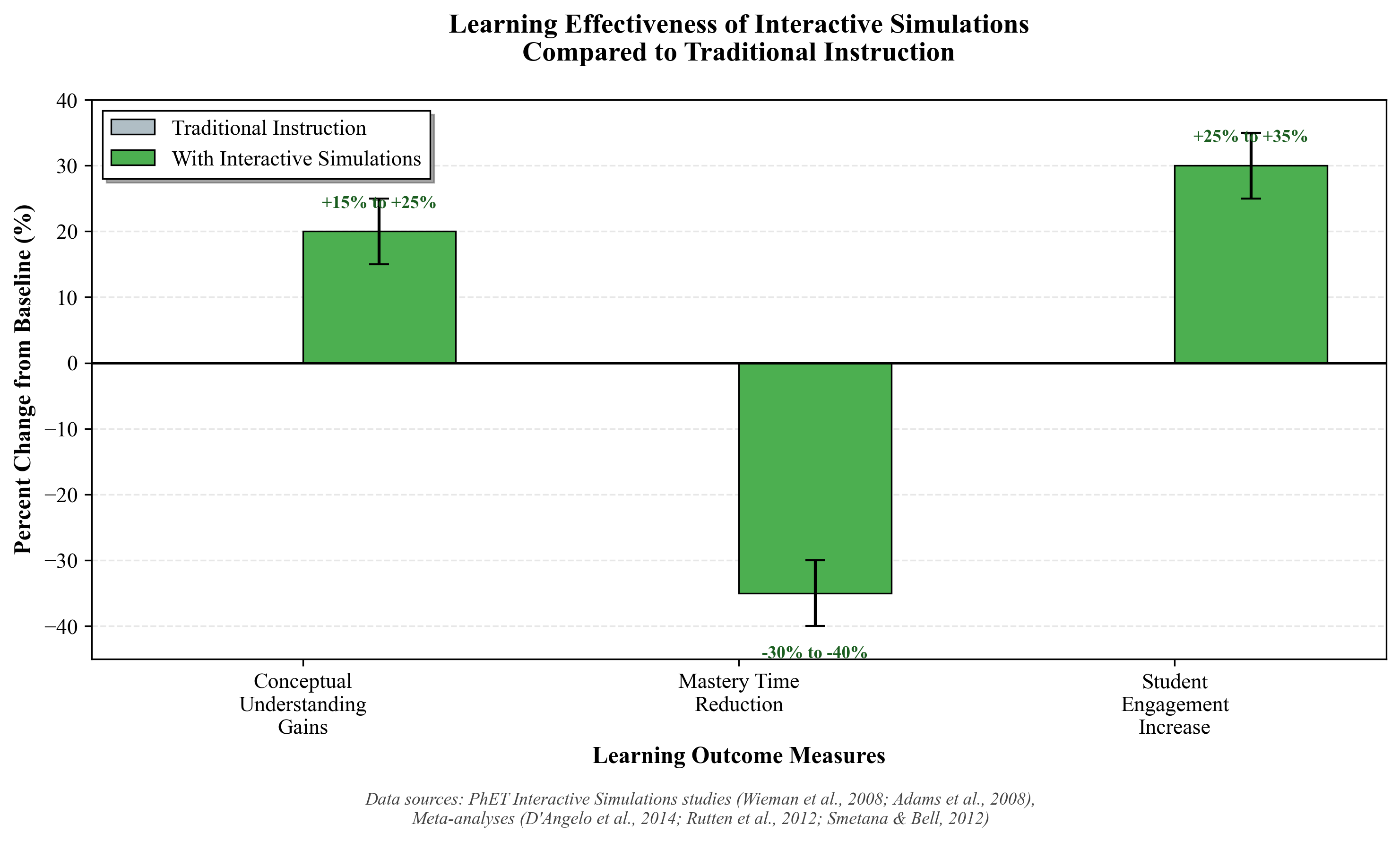}
\caption{Learning effectiveness of interactive simulations compared to traditional instruction across three key outcome measures. Data represent meta-analysis findings from multiple studies including PhET Interactive Simulations research and systematic reviews across STEM disciplines. Error bars indicate reported ranges of improvement. Sources: Wieman et al. (2008), Adams et al. (2008), D'Angelo et al. (2014), Rutten et al. (2012), Smetana \& Bell (2012).}
\label{fig:effectiveness}
\end{figure}

These impressive gains, visualized in Figure~\ref{fig:effectiveness}, reflect simulations' ability to make abstract concepts tangible and provide immediate, interactive feedback that supports active learning. By allowing learners to \textit{visualize invisible processes} such as magnetic fields, molecular motions, or electric currents, simulations bridge the gap between theoretical knowledge and real-world phenomena \cite{phet2023}.

\subsection{PhET Interactive Simulations: Evidence at Scale}

The PhET Interactive Simulations project at the University of Colorado Boulder provides the most comprehensive evidence for simulation effectiveness at scale. With over 45 million simulation runs annually across 175 countries, PhET represents one of the largest deployments of educational simulations worldwide \cite{phet2023}. This massive usage scale, combined with extensive research validation, demonstrates both the practical feasibility and educational effectiveness of well-designed interactive simulations.

Studies of PhET simulations have demonstrated significant learning gains across diverse populations and educational contexts \cite{adams2008study, finkelstein2005phet, perkins2006phet}. In controlled experiments comparing simulation-based instruction to traditional approaches, students using PhET simulations consistently achieved higher scores on conceptual assessments and demonstrated superior long-term retention.

\textbf{Controlled Experimental Evidence}: In a particularly compelling controlled study, high school physics students were randomly assigned to learn circuit concepts using either PhET Circuit Construction Kit simulations or physical circuit equipment (wires, bulbs, resistors). Despite equivalent instructional time and support, students using the simulation significantly outperformed their peers who used physical equipment on subsequent conceptual exam questions. Most remarkably, six weeks after the initial instruction, the simulation group demonstrated substantially better knowledge retention, suggesting that the interactive visual feedback provided by the simulation created more durable mental models \cite{finkelstein2005phet}.

\textbf{Transfer to Physical Skills}: Perhaps most surprising, when both groups were eventually asked to construct physical circuits with real equipment, students initially trained on simulations performed the task faster and with greater confidence than those who had originally used physical apparatus. This finding challenges the common assumption that virtual experiences cannot adequately prepare students for hands-on tasks, instead suggesting that well-designed simulations can provide superior conceptual foundations that transfer effectively to practical applications \cite{finkelstein2005phet}.

\textbf{Classroom Implementation Patterns}: Survey research with hundreds of PhET-using educators reveals that teachers employ simulations flexibly across diverse pedagogical contexts: demonstrations during lectures, guided inquiry laboratory activities, homework assignments, and remediation for struggling students. The primary goals educators report for using simulations include developing conceptual understanding, promoting scientific inquiry skills, and addressing known misconceptions \cite{phet2023}.

\subsection{Limitations and Challenges}

Despite their demonstrated effectiveness, educational simulations face several important limitations and challenges that must be acknowledged and addressed through ongoing research and development.

\textbf{Scope Constraints}: The deliberately focused nature of MicroSims means individual simulations cannot address all learning objectives within a domain. Complex, multi-faceted concepts may require sequences of simulations or complementary instructional approaches. Educators must thoughtfully integrate simulations within broader curriculum frameworks rather than viewing them as comprehensive, standalone educational solutions.

\textbf{Abstraction and Simplification}: By necessity, simulations simplify reality to make phenomena comprehensible and computable. While these simplifications often serve legitimate pedagogical purposes, they risk creating misconceptions if students do not understand model limitations. Simulations should explicitly document their assumptions and help students recognize that models are useful approximations rather than complete representations of reality.

\textbf{Technology Access and Digital Divide}: While browser-based simulations reduce technology barriers compared to specialized software, they still require reliable internet access and functional computing devices. Socioeconomic disparities in technology access risk making simulation-based learning another dimension of educational inequality unless intentionally addressed through school technology provisioning and alternative access strategies.

\textbf{Quality Variability}: Not all simulations prove equally effective. Poorly designed simulations can confuse rather than clarify, promote misconceptions, or simply fail to engage students productively. The demonstrated success of research-tested simulations like PhET underscores the importance of evidence-based design and iterative refinement. However, the MicroSims framework's goal of enabling rapid AI-assisted generation introduces quality assurance challenges requiring systematic approaches to validation and improvement.

\textbf{Teacher Preparation and Support}: Effective simulation use requires thoughtful pedagogical integration. Teachers need support in selecting appropriate simulations, designing accompanying activities with optimal scaffolding levels, and facilitating productive classroom discussions around simulation experiences. Professional development and high-quality instructional materials represent essential infrastructure for realizing simulations' educational potential.

\textbf{Assessment Integration}: Evaluating learning from simulation experiences presents ongoing challenges. While simulations generate rich interaction data potentially valuable for formative assessment, translating simulation performance into meaningful learning evaluation remains underdeveloped. Embedding meaningful assessment within simulations while maintaining their lightweight character requires continued innovation.

\subsection{Intelligent Textbook Integration Example}

The MicroSims framework envisions simulations not as isolated educational resources but as integrated components of intelligent, adaptive learning systems. An intelligent textbook system demonstrates how simulations can function within broader educational ecosystems to provide personalized, data-driven learning experiences.

\textbf{Adaptive Content Sequencing}: An intelligent textbook incorporating MicroSims continuously assesses student understanding through interaction patterns, simulation performance, and embedded assessment items. Based on this ongoing evaluation, the system adaptively sequences content---determining when students have sufficiently mastered prerequisites to advance, when additional practice through simulation exploration would benefit learning, or when remediation through alternative representations might address persistent misconceptions.

\textbf{Personalized Simulation Selection}: Rather than presenting identical content to all students, an intelligent textbook can select from libraries of related simulations targeting the same concepts at different complexity levels or through different contextual framings. A student struggling with abstract mathematical representations might receive a simulation emphasizing concrete, everyday contexts, while an advanced student might encounter more sophisticated parameter spaces and quantitative analysis tasks.

\textbf{Learning Analytics and Feedback Loops}: Comprehensive logging of student interactions with embedded simulations---including time spent, parameters explored, challenges attempted, and errors made---provides rich data for learning analytics. Intelligent systems analyze these data streams to identify struggling students requiring intervention, recognize common misconceptions requiring instructional attention, and refine content recommendations based on observed effectiveness patterns.

\textbf{Integration with Reinforcement Learning}: Advanced implementations might employ reinforcement learning algorithms that treat simulation selection and sequencing as optimization problems: the system learns through trial and error which combinations of simulations, scaffolding levels, and timing produce optimal learning outcomes for different student profiles. Over time, such systems become increasingly effective at personalizing educational experiences as they accumulate data about what works for whom under what circumstances.

This vision of intelligent textbook integration positions MicroSims as essential building blocks for next-generation adaptive learning systems---systems that combine the engagement and interactivity of simulations with the personalization and continuous improvement capabilities of AI-driven education technology.

\section{Discussion}
\label{sec:discussion}

\subsection{Implications for Educational Equity}

The unique characteristics of MicroSims have significant implications for educational equity:

\textbf{Reduced Cost Barriers}: Traditional educational software often requires expensive licenses, powerful hardware, or high-bandwidth internet connections. MicroSims, being lightweight and self-contained, can run on basic devices with minimal connectivity, making quality interactive content accessible to under-resourced schools and students.

\textbf{Language and Cultural Adaptation}: AI systems can generate MicroSims in different languages or adapt them for different cultural contexts on demand, without requiring separate development efforts for each market.

\textbf{Accessibility by Design}: Standardized patterns include accessibility features, ensuring that generated MicroSims support screen readers, keyboard navigation, and other assistive technologies.

\subsection{Network Effects and Standardization}

The standardized architecture of MicroSims creates powerful network effects as adoption increases. Each new MicroSim created following the framework patterns contributes to a growing ecosystem that benefits all users.

\subsection{Limitations and Challenges}

Despite their advantages, MicroSims face several limitations that warrant discussion:

\textbf{Scope Constraints}: The deliberately focused nature of MicroSims means they cannot address all learning objectives. Complex, multi-faceted concepts may require sequences of MicroSims or complementary instructional approaches.

\textbf{AI Generation Quality}: While generative AI has made remarkable progress, the quality of AI-generated MicroSims still requires human review and refinement to ensure pedagogical soundness and functional correctness.

\textbf{Assessment Integration}: Embedding meaningful assessment within MicroSims while maintaining their lightweight character presents ongoing challenges.

\subsection{Broader Impact}

The MicroSim framework has implications beyond individual simulation creation, potentially influencing how educational technology is conceptualized, developed, and deployed more broadly.

\section{Conclusion}
\label{sec:conclusion}

This paper has introduced MicroSims, a comprehensive framework for creating lightweight, interactive educational simulations that address persistent barriers to widespread adoption of simulation-based learning. By occupying the unique intersection of simplicity, accessibility, and AI-generation capability, MicroSims enable educators worldwide to create custom, curriculum-aligned simulations on demand.

\subsection{Summary of Contributions}

We have presented:
\begin{enumerate}
\item A comprehensive design framework encompassing technical architecture, pedagogical principles, and user experience guidelines
\item Evidence that standardized patterns enable reliable AI-assisted generation
\item An iframe-based distribution model providing universal embedding across learning platforms
\item A metadata framework supporting discovery, personalization, and learning analytics
\item Empirical evidence from educational research demonstrating simulation effectiveness
\end{enumerate}

\subsection{Future Directions}

Several promising research directions emerge from this work:

\textbf{Enhanced AI Capabilities}: As language models continue to improve, future work should explore more sophisticated simulation generation, including adaptive difficulty adjustment and personalized content creation based on individual student performance data.

\textbf{Learning Analytics Integration}: Deeper integration with learning analytics systems could enable real-time adaptation of simulation parameters based on aggregate student interaction patterns.

\textbf{Collaborative Simulations}: Extending the framework to support multi-user, collaborative simulation experiences while maintaining the lightweight architecture.

\textbf{Immersive Technologies}: Exploring integration with virtual and augmented reality platforms while preserving the core principles of simplicity and accessibility.

\subsection{Conclusion}

MicroSims represent a paradigm shift in educational content creation, transforming simulation development from a specialized, resource-intensive process to an accessible, AI-assisted workflow available to any educator. By removing traditional barriers of cost, technical complexity, and platform dependence, MicroSims democratize access to interactive educational experiences. As generative AI continues to advance, the MicroSim framework provides a foundation for the next generation of adaptive, personalized learning systems.

The ultimate goal remains unchanged: creating learning experiences that inspire curiosity, build understanding, and empower students to apply knowledge in meaningful ways. MicroSims represent one step toward realizing the vision of truly adaptive, universally accessible educational technology that serves all learners effectively.

\section*{Acknowledgments}
The author would like to thank the open-source p5.js community, the educational technology research community, and the contributors to the PhET Interactive Simulations project whose research has informed this work.



\begin{thebibliography}{99}

\bibitem{stephenson1995}
Stephenson, Neal.
\textit{The Diamond Age: Or, A Young Lady's Illustrated Primer}.
Bantam Spectra, New York, NY, 1995.

\bibitem{wieman2008phet}
Wieman, Carl E., Adams, Wendy K., and Perkins, Katherine K.
PhET: Simulations that enhance learning.
\textit{Science}, 322(5902):682--683, 2008.

\bibitem{adams2008study}
Adams, Wendy K., Reid, Sam, LeMaster, Ron, McKagan, Sarah B., Perkins, Katherine K., Dubson, Michael, and Wieman, Carl E.
A study of educational simulations part I - Engagement and learning.
\textit{Journal of Interactive Learning Research}, 19(3):397--419, 2008.

\bibitem{finkelstein2005phet}
Finkelstein, Noah D., Adams, Wendy K., Keller, Christine J., Kohl, Patrick B., Perkins, Katherine K., Podolefsky, Noah S., Reid, Sam, and LeMaster, Ron.
When learning about the real world is better done virtually: A study of substituting computer simulations for laboratory equipment.
\textit{Physical Review Special Topics - Physics Education Research}, 1(1):010103, 2005.

\bibitem{perkins2006phet}
Perkins, Katherine, Adams, Wendy, Dubson, Michael, Finkelstein, Noah, Reid, Sam, Wieman, Carl, and LeMaster, Ron.
PhET: Interactive simulations for teaching and learning physics.
\textit{The Physics Teacher}, 44(1):18--23, 2006.

\bibitem{phet2023}
University of Colorado Boulder.
PhET Interactive Simulations, 2023.
\url{https://phet.colorado.edu/}. Accessed: 2024-01-15.

\bibitem{microsims2024}
McCreary, Dan, Lockhart, Valerie, and Peterson, Troy A.
MicroSims: Educational Simulations Library, 2024.
\url{https://dmccreary.github.io/microsims/}.
Collection of 100+ interactive educational simulations demonstrating the framework principles.

\bibitem{dangelo2014simulations}
D'Angelo, Cynthia, Rutstein, Daisy, Harris, Christopher, Bernard, Robert, Borokhovski, Eugen, and Haertel, Geneva.
Simulations for STEM learning: Systematic review and meta-analysis.
Technical report, SRI International, 2014.

\bibitem{merchant2014effectiveness}
Merchant, Zahira, Goetz, Ernest T., Cifuentes, Lauren, Keeney-Kennicutt, Wheijen, and Davis, Trina J.
Effectiveness of virtual reality-based instruction on students' learning outcomes in K-12 and higher education: A meta-analysis.
\textit{Computers \& Education}, 70:29--40, 2014.

\bibitem{rutten2012learning}
Rutten, Nico, van Joolingen, Wouter R., and van der Veen, Jan T.
The learning effects of computer simulations in science education.
\textit{Computers \& Education}, 58(1):136--153, 2012.

\bibitem{smetana2012computer}
Smetana, Lara K. and Bell, Randy L.
Computer simulations to support science instruction and learning: A critical review of the literature.
\textit{International Journal of Science Education}, 34(9):1337--1370, 2012.

\bibitem{freeman2014active}
Freeman, Scott, Eddy, Sarah L., McDonough, Miles, Smith, Michelle K., Okoroafor, Nnadozie, Jordt, Hannah, and Wenderoth, Mary Pat.
Active learning increases student performance in science, engineering, and mathematics.
\textit{Proceedings of the National Academy of Sciences}, 111(23):8410--8415, 2014.

\bibitem{prince2004active}
Prince, Michael.
Does active learning work? A review of the research.
\textit{Journal of Engineering Education}, 93(3):223--231, 2004.

\bibitem{sweller1988cognitive}
Sweller, John.
Cognitive load during problem solving: Effects on learning.
\textit{Cognitive Science}, 12(2):257--285, 1988.

\bibitem{zaharias2009usability}
Zaharias, Panagiotis and Poylymenakou, Angeliki.
Developing a usability evaluation method for e-learning applications: Beyond functional usability.
\textit{International Journal of Human-Computer Interaction}, 25(1):75--98, 2009.

\bibitem{davids2015usability}
Davids, Mitchell R., Chikte, Usuf M. E., and Halperin, Mitchell L.
Effect of improving the usability of an e-learning resource: A randomized trial.
\textit{Advances in Physiology Education}, 39(2):59--66, 2015.

\bibitem{wilensky1999netlogo}
Wilensky, Uri.
NetLogo.
Center for Connected Learning and Computer-Based Modeling, Northwestern University, Evanston, IL, 1999.
\url{http://ccl.northwestern.edu/netlogo/}.

\bibitem{lms2023}
Learning Management Systems in Higher Education.
Industry report on LMS adoption and usage, 2023.

\bibitem{xapi2024}
Advanced Distributed Learning (ADL) Initiative.
Experience API (xAPI) Specification, 2024.
\url{https://github.com/adlnet/xAPI-Spec}.
Version 1.0.3. Learning technology specification for tracking and analyzing learning experiences.

\bibitem{khanacademy2023}
Khan Academy.
Khanmigo: AI-powered learning guide, 2023.
\url{https://www.khanacademy.org/khan-labs}. Accessed: 2024-01-15.

\bibitem{eric2009}
Hung, William.
An Investigation of the Potential of Interactive Simulations for Developing System Thinking Skills in Elementary School: A Case Study with Fifth-Graders and Sixth-Graders.
\textit{International Journal of Science Education}, 31(9):1183--1208, March 2009.

\bibitem{mdpi2024}
Rosário, Ana Teresa, Cardoso, Ana Paula, and Rodrigues, Paulo.
Digital Simulations in STEM Education: Insights from Recent Empirical Studies, a Systematic Review.
\textit{Digital}, 5(1):10, 2024.
MDPI.

\end{thebibliography}
\end{document}